\documentclass{article}

\usepackage{PRIMEarxiv}

\usepackage[utf8]{inputenc} 
\usepackage[T1]{fontenc}    
\usepackage{hyperref}       
\usepackage{url}            
\usepackage{booktabs}       
\usepackage{amsfonts}       
\usepackage{nicefrac}       
\usepackage{microtype}      
\usepackage{lipsum}
\usepackage{fancyhdr}       
\usepackage{graphicx}       
\graphicspath{{media/}}     
\usepackage{rotating}
\usepackage{graphicx}
\usepackage{xcolor}
\usepackage{lscape}
\usepackage{multirow}
\usepackage{float}
\usepackage{tabularray}
\usepackage{adjustbox}
\usepackage{tabularx,ragged2e,booktabs}
\usepackage{tabulary,booktabs}
\usepackage{tabularray}
\title{\textit{hist2RNA}: An efficient deep learning architecture to predict gene expression from breast cancer histopathology images
}

\author{
  Raktim Kumar Mondol \\
    School of Computer Science and Engineering \\
  UNSW Sydney \\
  Kensington, NSW 2052, Australia
   \And
  Ewan K.A. Millar\\
  Department of Anatomical Pathology,  \\
  NSW Health Pathology, St. George Hospital,  \\
  Kogarah, NSW 2217, Australia
  \AND
  Peter H Graham, Lois Browne \\
  Cancer Care Centre, St George Hospital \\
  Kogarah, NSW 2217, Australia 
  \AND
  Arcot Sowmya, Erik Meijering \\
  School of Computer Science and Engineering \\
  UNSW Sydney \\
  Kensington, NSW 2052, Australia\\
  \texttt{\{a.sowmya, erik.meijering\}@unsw.edu.au}
}

\begin{document}
\maketitle
\begin{abstract}
Gene expression can be used to subtype breast cancer with improved prediction of risk of recurrence and treatment responsiveness over that obtained using routine immunohistochemistry (IHC). However, in the clinic, molecular profiling is primarily used for ER+ breast cancer, which is costly, tissue destructive, requires specialised platforms, and takes several weeks to obtain a result. Deep learning algorithms can effectively extract morphological patterns in digital histopathology images to predict molecular phenotypes quickly and cost-effectively. We propose a new, computationally efficient approach called hist2RNA inspired by bulk RNA sequencing techniques to predict the expression of 138 genes (incorporated from 6 commercially available molecular profiling tests), including luminal PAM50 subtype, from hematoxylin and eosin (H\&E)-stained whole slide images (WSIs). The training phase involves the aggregation of extracted features for each patient from a pretrained model to predict gene expression at the patient level using annotated H\&E images from The Cancer Genome Atlas (TCGA, n
 = 335). We demonstrate successful gene prediction on a held-out test set (n = 160, corr = 0.82 across patients, corr = 0.29 across genes) and perform exploratory analysis on an external tissue microarray (TMA) dataset (n = 498) with known IHC and survival information. Our model is able to predict gene expression and luminal PAM50 subtype (Luminal A versus Luminal B) on the TMA dataset with prognostic significance for overall survival in univariate analysis (c-index = 0.56, hazard ratio = 2.16 (95\% CI 1.12--3.06), \emph{p} < 5 $\times$ 10\textsuperscript{$-$3}), and independent significance in multivariate analysis incorporating standard clinicopathological variables (c-index = 0.65, hazard ratio = 1.87 (95\% CI 1.30--2.68), \emph{p} < 5 $\times$ 10\textsuperscript{$-$3}). The proposed strategy achieves superior performance while requiring less training time, resulting in less energy consumption and computational cost compared to patch-based models. Additionally, hist2RNA predicts gene expression that has potential to determine luminal molecular subtypes which correlates with overall survival, without the need for expensive molecular testing.
\end{abstract}




\section{Introduction}
Breast cancer is one of the most common cancers worldwide and a leading cause of death in women \cite{Wilkinson2022}. In Australia, over 3,000 women die each year from it, although the 5 year survival has increased from 76\% to 91\% , which has been achieved through early diagnosis and improved treatment regimes \cite{bcdata}\cite{Ji2020}. Histopathology, clinical findings and imaging (CT, MRI, Ultrasound) are used for breast cancer diagnosis and stage classification \cite{Hirra2021}\cite{Conti2021}. Hematoxylin and eosin (H\&E) histopathology slides contain essential phenotypic information and are used to provide key diagnostic and prognostic features (size, type, grade, lymph nodal status and expression of the biomarkers ER, PR, HER2,  Ki67) to guide treatment decisions \cite{Zhou2020}.

A critical clinical question is the selection of the postmenopausal luminal ER+ patients who may benefit from the addition of chemotherapy to endocrine therapy \cite{Paik2006}\cite{Albain2010}\cite{Sparano2018}. This difficult decision may be further supported by commercially available multigene tests to assess the risk of recurrence (e.g.\ Oncotype DX, Prosigna, Mammaprint, Endopredict etc) and molecular subtype (PAM50: Luminal A, Luminal B, HER2-enriched, or Basal). However, molecular profiling is expensive, tissue destructive, takes several weeks of obtain a result, may not be reimbursed by healthcare providers and therefore not available to all patients with breast cancer \cite{Ki2020}\cite{deverka_dreyfus_2014}. Providing a cheap and scalable solution to predict molecular subtype using deep learning could make this important clinical information rapidly available to more patients. Whilst simplified IHC based intrinsic molecular subtyping can be used as a surrogate for gene expression assays, it is less robust \cite{Szymiczek2021}. Therefore, the use of a rapid and low cost method for gene expression prediction from digital images could offer clinically useful information for more patients.

With the advent of phenotype-rich medical images and next-generation sequencing (NGS) technology, artificial intelligence (AI) has led to the development of newer computer aided diagnosis (CAD) systems capable of recognising complex patterns to aid in clinical decision making \cite{Lu2021}\cite{Karthiga2021}. Deep learning (DL), a subset of AI, has demonstrated enormous potential for analysing pathology images and gene expression data, driving towards the goal of personalised treatment and identifying the underlying key molecular features of various diseases \cite{Yu2021}\cite{Katzman2018}\cite{Ragab2022}. Deep learning models can also be used for molecular phenotyping (gene mutations, gene transcripts and proteins), thereby enabling translational research that may eventually facilitate detailed characterisation of individual tumours and inform clinically targeted treatment decisions \cite{Qu2021}\cite{Sammut2022}\cite{Binder2021}. Intervariability in histology slide staining is a challenge, however various techniques can be applied to reduce bias and undesirable colour variation \cite{Macenko2009}\cite{Vahadane2016}. In addition, data augmentation \cite{Sandfort2019}, multiresolution patches \cite{Hong2021} and overlapping patches \cite{Bernal2019} can be used to improve model generalisation and performance.

Several studies have proposed different approaches to predict molecular phenotypes from whole slide images (WSIs). Hong et al proposed the use of multiple patches of different resolutions to enable a neural network to learn features at different scales, whereas Schmauch et al used a clustering approach to reduce the dimensions and give a final prediction per WSI. Various pretrained models, such as ResNet18, ResNet50, ShuffleNet and InceptionV3, have been successfully used for molecular phenotype prediction in recent years \cite{Hong2021}\cite{Schmauch2020} \cite{Schrammen2021}\cite{Weitza}\cite{Wang2021f}. Notably, ShuffleNet is particularly efficient due to its use of depth-wise separable convolution \cite{Hoang2019}\cite{Zhang2018}. In addition, Tavolara et al employed a custom convolutional neural network (CNN) that utilises multiple instance learning (MIL) based attention pooling to predict gene expression, eliminating the need for expert annotation.~\cite{Tavolara2021}. In our own research, we have experimented with several state-of-the-art pretrained deep learning models such as ResNet50, InceptionV3, RegNetY320, EfficientNetV2S and DenseNet201 for the prediction of molecular phenotypes from WSIs, and found that existing models typically require excessive training time and computational resources.

In this paper, we propose a new deep learning method for gene expression prediction from breast cancer histopathology images that requires substantially less training time, computational resources and energy usage, without impacting prediction performance. Our proposed hist2RNA model is inspired by bulk RNA sequencing that relies on averaged gene expression from a homogenised population of cells \cite{Hegenbarth2022}. Bulk RNA sequencing, however, has some drawbacks because it cannot show the specific composition of each cell type or the heterogeneity of the tissue. Our experimental results show that it is nevertheless possible to use histopathology images to efficiently predict gene expression to gain insight into disease biology and to better predict cancer subtypes for targetted treatment.

\section{Materials and Methods}

\subsection{Study Overview}
In this study, we aimed to predict mRNA expression levels of 138 genes using deep CNN model on WSIs. To prepare the WSIs for analysis, we first annotated the cancer region, extracted image patches, and applied colour normalisation. These preprocessed images were subsequently utilized to train and optimize the deep CNN model by tuning hyperparameters and adjusting the model architecture. Finally, the optimized models were evaluated on a held-out test set and an external dataset.
\subsection{Data Collection}
Our study consisted of H\&E-stained formalin-fixed paraffin-embedded (FFPE) digital slides from TCGA-BRCA (n=495). We selected a sample of 495 cases out of total 1,100 cases due to the complex and labor-intensive nature of annotating histopathological images. In order to maintain high data quality, our expert pathologist carefully reviewed the slides, leading to a smaller but more reliable dataset. WSIs from TCGA-BRCA were downloaded from the \href{https://portal.gdc.cancer.gov/}{GDC Portal} (Accessed August 20, 2021). We used samples from all four molecular subtypes for our experiment: Luminal A (LumA, 174 samples), Luminal B (LumB, 116 samples), Basal (135 samples), and HER2 (70 samples). All patients had corresponding RNA-sequencing data available for analysis. For our study, we used 335 samples for training and 160 samples for testing (held-out test set) while maintain the relative proportions of all subtypes. We also used an external TMA dataset of 498 patients with invasive breast cancer from a randomised radiotherapy clinical trial at St George Hospital, Sydney, Australia (ClinicalTrials.gov NCT00138814) (Accessed January 23, 2022) \cite{Millar2020}. Ethics approval was provided by South East Sydney Local Health District Human Research Ethics Committee at the Prince of Wales Hospital, Sydney (HREC 96/16). Samples of the 498 patients were scattered randomly across 18 glass slides. Each slide contained multiple cores from different patients, and each patient's tumour was sampled with a $3\times1$ mm core using the Beecher Manual Arrayer MTA-1 as previously published \cite{Millar2020}. This dataset included a range of clinical information for each patient, such as follow-up duration, overall survival status, tumor grade, tumor size, lymph node status, age at diagnosis, and intrinsic subtypes determined by IHC. We tested model performance to predict survival on this external TMA dataset with additional exploratory analysis.

\subsection{Data Preparation}
\subsubsection{Slide Annotation}
An expert breast pathologist manually annotated the selected slides using QuPath \cite{Bankhead2017}. The annotation was performed for localisation of the tumour outline, excluding any necrosis but including stroma and tumour infiltrating lymphocytes (TILs). The pathologist was blinded to any molecular or clinical features during annotation.

\subsubsection{Image Data Preprocessing}
\begin{figure}[!t]
\centering
\includegraphics[width=0.9\textwidth]{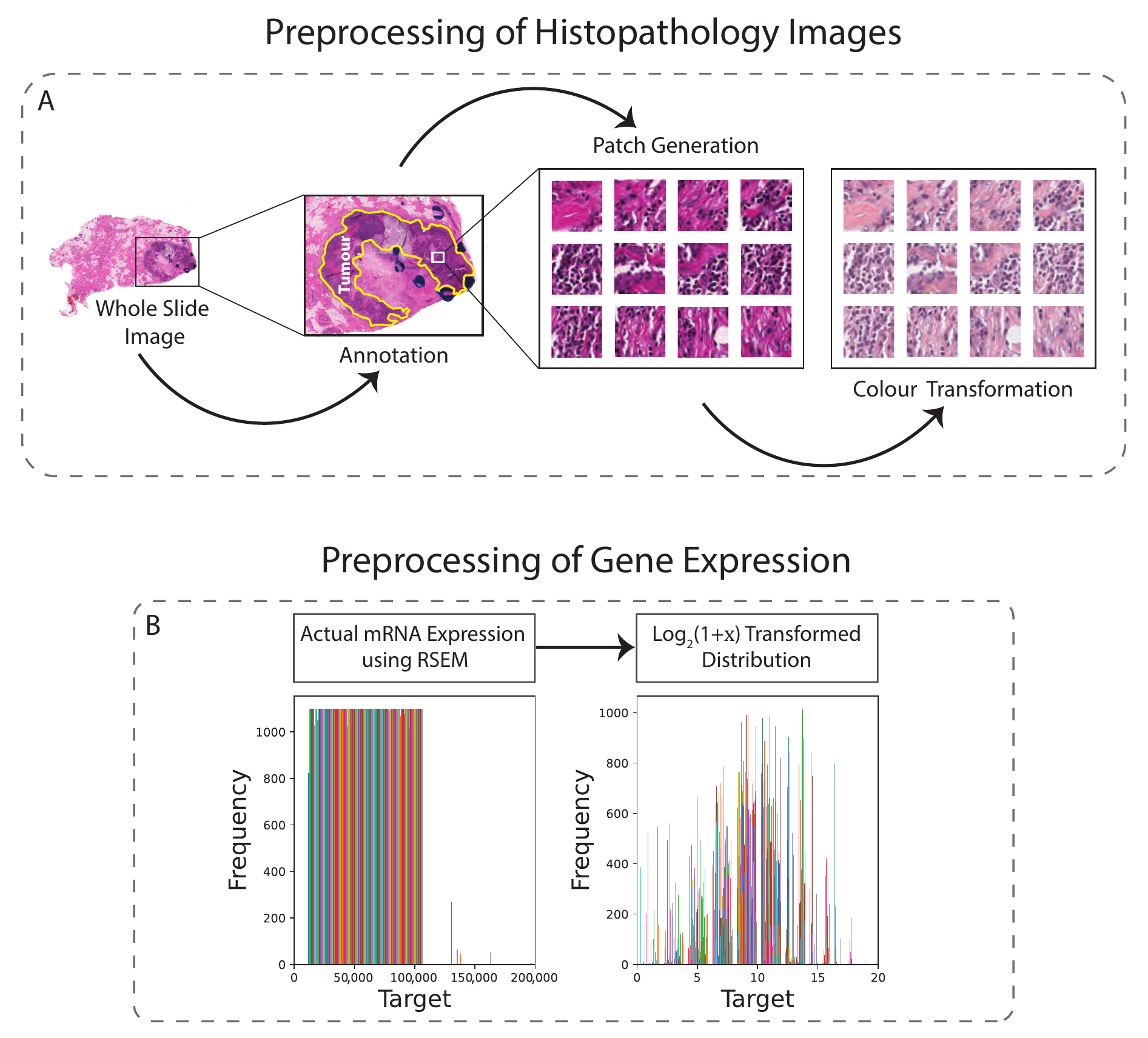}
\caption{Preprocessing technique for histopathology images and gene expression. A: WSIs are first annotated, then non-overlapping patches are generated from the annotated region, and finally colour transformation is applied. B: Gene values are $\log_2(1+x)$ normalised.}
\label{fig:preprocessing}
\end{figure}

In this study, H\&E-stained tissues were taken from both the TCGA and TMA datasets. In order to optimize the balance between image resolution and file size, both datasets were downsampled to 0.25 µm/pixel (approximately ×40 magnification). A semi-automated approach using QuPath software was used to generate tissue masks, which excluded areas with artifacts and non-tissue regions. The annotated tumour regions were then tiled into $224\times224$ pixel patches, yielding approximately 1,000 non-overlapping patches per sample. To address staining inconsistencies, vector-based colour normalisation was applied to both datasets, resulting in improved quantitative results~\cite{Macenko2009} (Figure~\ref{fig:preprocessing}A). Overall, the preprocessing steps for both datasets were conducted with the goal of producing high-quality image patches suitable for subsequent analysis.
\subsubsection{RNA-Sequencing Data Preprocessing}
We obtained transcriptome-wide RNA-sequencing data representing mRNA expression levels for a total of 20,438 genes in the reference genome from the TCGA dataset, which were processed using RNA-sequencing by expectation maximization (RSEM) \cite{Li2011}, and downloaded from the \href{https://www.cbioportal.org/}{cBioPortal} platform. Only patients with both RNA-sequencing data and WSIs available were included in the study. We extracted genes used in the following commercially available assays: Oncotype DX, Mammaprint,Prosigna (PAM50), EndoPredict, BCI (Breast Cancer Index), and Mammostrat, which resulted in 138 genes as the final training targets (see Supplementary Materials for gene lists).

Besides images, the genetic data also required preprocessing because regression analysis directly on raw RNA-sequencing data would lead the model to focus only on the most strongly expressed genes, which would lead to poor model performance. As the normalised gene counts ($x$) can be equal to zero, we shifted them by 1 before log-transformation. Thus, we applied $\log_2(1 + x)$ transformation on the mRNA expression values (Figure~\ref{fig:preprocessing}B). This transformed data is typically less skewed, with fewer extreme values, but may contain unequal variances for the covariates \cite{Zwiener2014}. Despite this, it is unlikely to be a major concern when using deep learning-based regression models, as these models can effectively adapt to unequal variances and other nonlinearities in the data, allowing the capture of complex relationships and patterns.

\begin{figure}[!t]
\includegraphics[width=16.8cm]{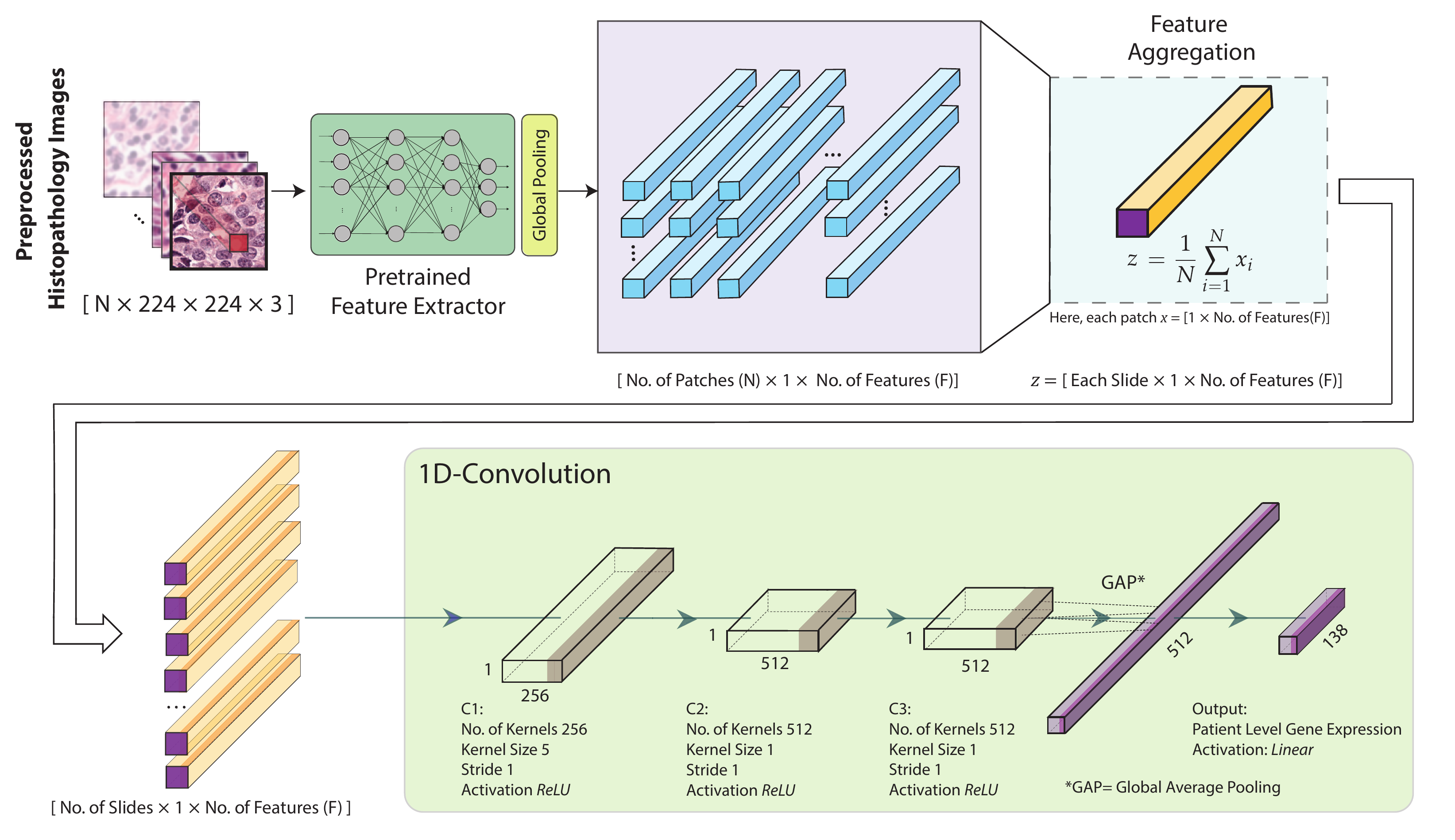}
\caption{Proposed hist2RNA neural network architecture. After image preprocessing, a pretrained model extracts features from the histopathology image patches. The extracted features are then aggregated for each patient. Next, three convolutional blocks (C1, C2, C3) are used, each consisting of a 1D convolutional layer with a rectified linear unit (ReLU) activation function. C1 uses 256 kernels of size 5 $\times$ 5, while C2 and C3 use 512 kernels of size 1 $\times$ 1. Finally, the output layer consists of 138 neurons with a linear activation function, followed by a global average pooling layer.}
\label{fig:architecture}
\end{figure}  

\subsection{Feature Extraction, Aggregation and Model Training}
The architecture of the proposed hist2RNA (Figure~\ref{fig:architecture}) uses a pretrained neural network to extract features from the image patches. More specifically, we considered five different pretrained models (EfficientNet, RegNet, DenseNet, Inception, ResNet) and evaluated their performance with our model. Feature extraction results in a tensor of dimension $N$ $\times$ $1$ $\times$ $F$, where $N$ is the number of patches and F the number of features. Our model aggregates the patch-level features to obtain a slide-level feature representation in the form of a single vector of dimension $1$ $\times$ $1$ $\times$ $F$ for each slide, as follows:
\begin{equation}\label{eq.1}
z=\frac{1}{N} \sum_{i=1}^{N} x_i
\end{equation}
where $z$ is the aggregated feature and each image patch is represented by a feature vector of size $1$ $\times$ $F$, denoted as $x_i$, where $i$ is the index of the patch. That is, the resulting slide-level vector has the same number of features as the original patch-level vectors, and represents their average. This aggregated feature is fed to three 1D convolutional layers followed by global average pooling and an output layer with 138 neurons with linear activation functions to build a regression model.

We employed the Adam optimizer with the mean-squared error loss function and learning rate of 0.001. We used a minibatch of 12 samples per step and early stopping to monitor change in loss with a patience of 4. We ran continuous optimisation until the early stopping criterion was met or the maximum of 150 epochs were completed. The hyperparameters used in the optimisation process were determined empirically. During optimisation we stored the best weights having the lowest mean-squared error loss for use in testing. Depending on when the early stopping criterion was met, optimisation runs took approximately 20 to 50 minutes on a single GPU (NVIDIA Tesla V100 32GB).

\subsection{Evaluation Metrics}
Predictions of gene expression and genetic variation (such as mutation status, cancer grade, chromosomal instability, copy number variation and methylation status) are, respectively, regression and classification problems. The performance of the regression models in this project was evaluated using the Spearman rank correlation coefficient and its corresponding FDR (False Discovery Rate) adjusted p-value. This p-value was calculated using a t-distribution-based approach to determine the statistical significance of the observed correlation between predicted and actual gene expression values. The coefficient of determination (R$^2$) was also employed as a metric to assess the effectiveness of our regression model. The R$^2$ score measures the proportion of the variance in the dependent variable that can be explained by the independent variable, with values typically ranging from 0 to 1, with higher scores indicating that the regression model has a better fit to the data. However, in cases where the model performs poorly, R$^2$ can be negative. These parameters are essential in evaluating the effectiveness of the regression problem. In addition to the analyses, t-tests were conducted to evaluate the differences between positive and negative groups for each biomarker (ER, PR, and HER2), while ANOVA tests were used to assess the variance in tumor grades (1, 2, 3). These statistical tests are instrumental in determining the presence of statistically significant differences among various groups.
\begin{figure}[!t]
\centering
\includegraphics[width=0.8\textwidth]{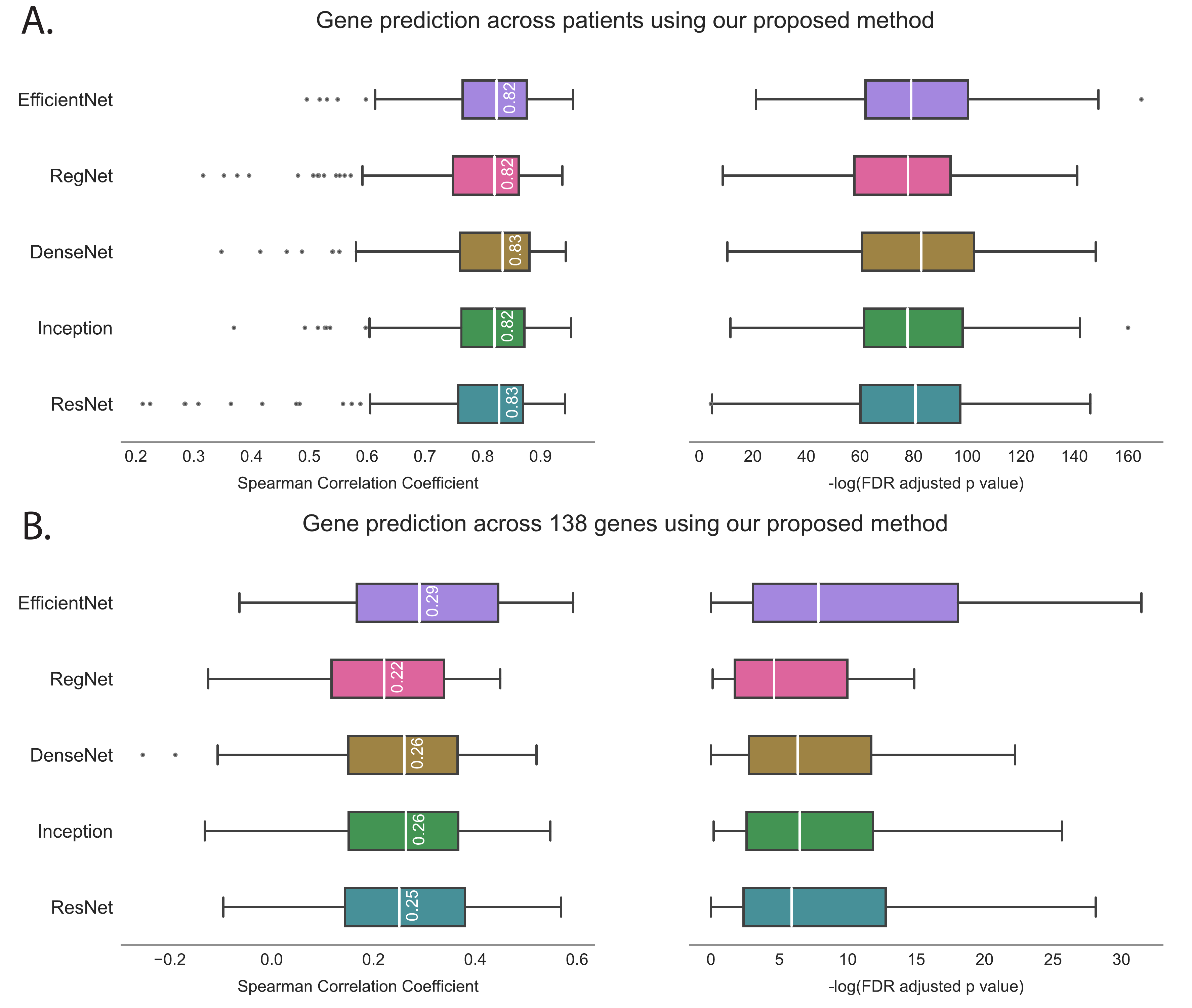}
\caption{Performance comparison of five pretrained models on the held-out test set using the proposed hist2RNA method. The box plots show the distribution of Spearman correlation coefficients and their corresponding $-$log FDR-adjusted p values (where significance is represented by $-$log(5\% FDR p value) > 1.30). The vertical white bar inside each box represents the median value, while the box itself indicates the interquartile range (IQR) from the 25th to the 75th percentile. The whiskers indicate the range of values beyond IQR and any data point that falls outside of the whiskers is considered an outlier and is marked with a gray circle. \textbf{A:} Gene prediction is performed across patients, with the correlation coefficient calculated using predicted and true expression levels of all 138 genes for each patient. \textbf{B:} Gene prediction is performed across 138 genes, with the correlation coefficient calculated for each gene separately, based on the predicted and true expression levels for all patients.}
\label{box_plot}
\end{figure}
Further, survival analysis was evaluated with the concordance index (c-index), which was computed with the lifelines package \cite{lifeline}. The c-index measures the effectiveness of predicted risk scores on ranking survival times, with 1 indicating perfect concordance and 0.5 representing results from random predictions. Both univariate and multivariate survival models were employed to investigate the relationship between survival information and various predictor variables, such as cancer subtypes, tumour grade, tumour size, age, and lymph node status. The strength of these relationships was assessed using hazard ratios (HRs), which were accompanied by their corresponding confidence intervals (CIs). HR represents the hazard or risk of an event (such as death) occurring in one group relative to another group, while controlling for other variables. In the context of multivariate analysis, the HR for a LumB cancer subtype compared to a LumA subtype would indicate the relative risk of death in LumB patients compared to LumA patients, while controlling for the effects of other factors that may influence survival outcomes such as age, tumour grade and lymph node status. On the other hand, in the univariate analysis, the HR for LumB subtype compared to LumA was calculated without adjusting for other factors, providing a simple assessment of the relationship between cancer subtypes and survival outcomes.

\begin{figure}[!t]
\centering
\includegraphics[width=0.8\textwidth]{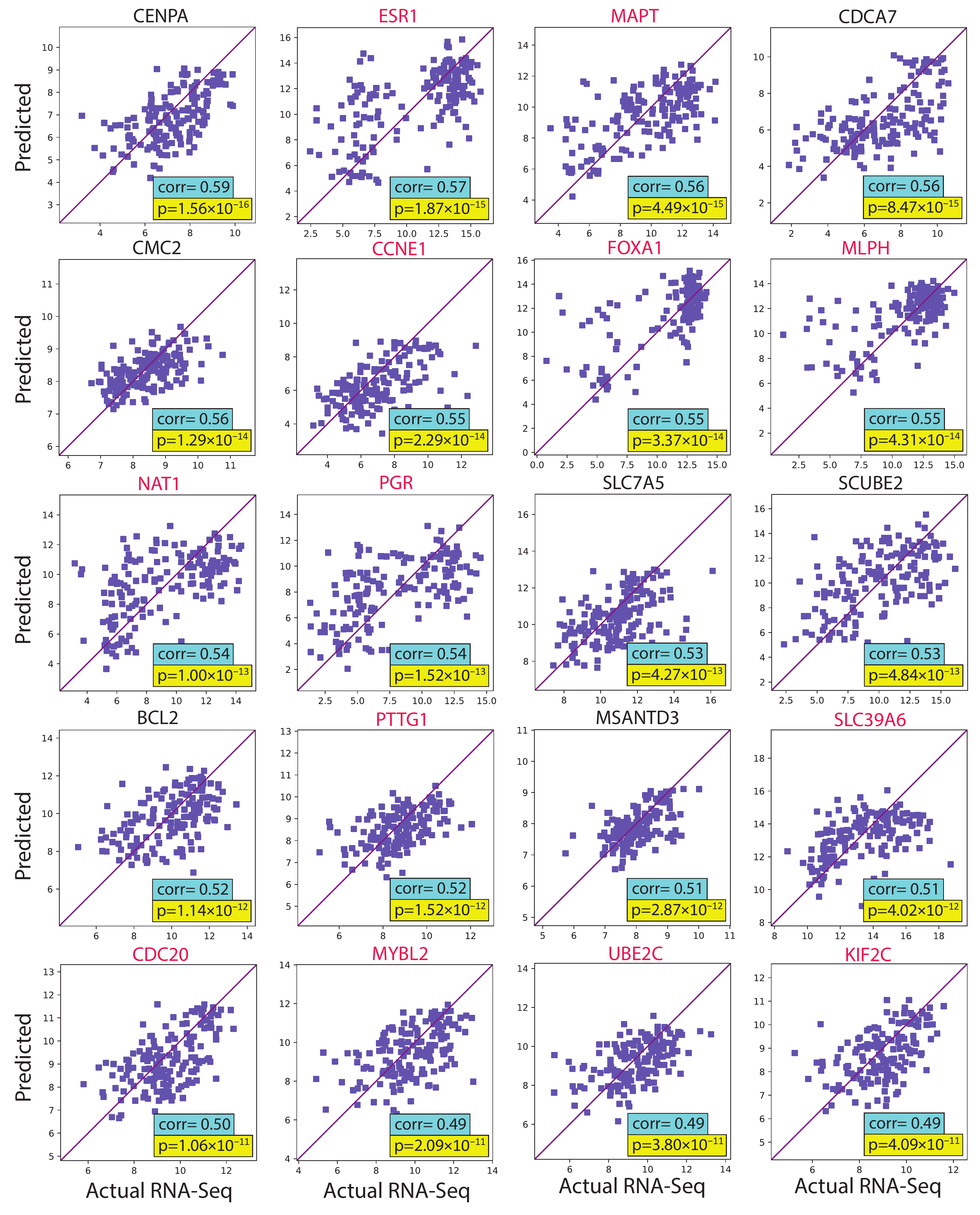}
\caption{Scatter plots of predicted and actual RNA-sequencing gene expression values for the 20 top genes ranked by correlation coefficient in the test set. PAM50 genes are highlighted in red.}
\label{scatter_plot}
\end{figure}

\begin{figure}[!t]
\centering
\includegraphics[width=0.65\textwidth]{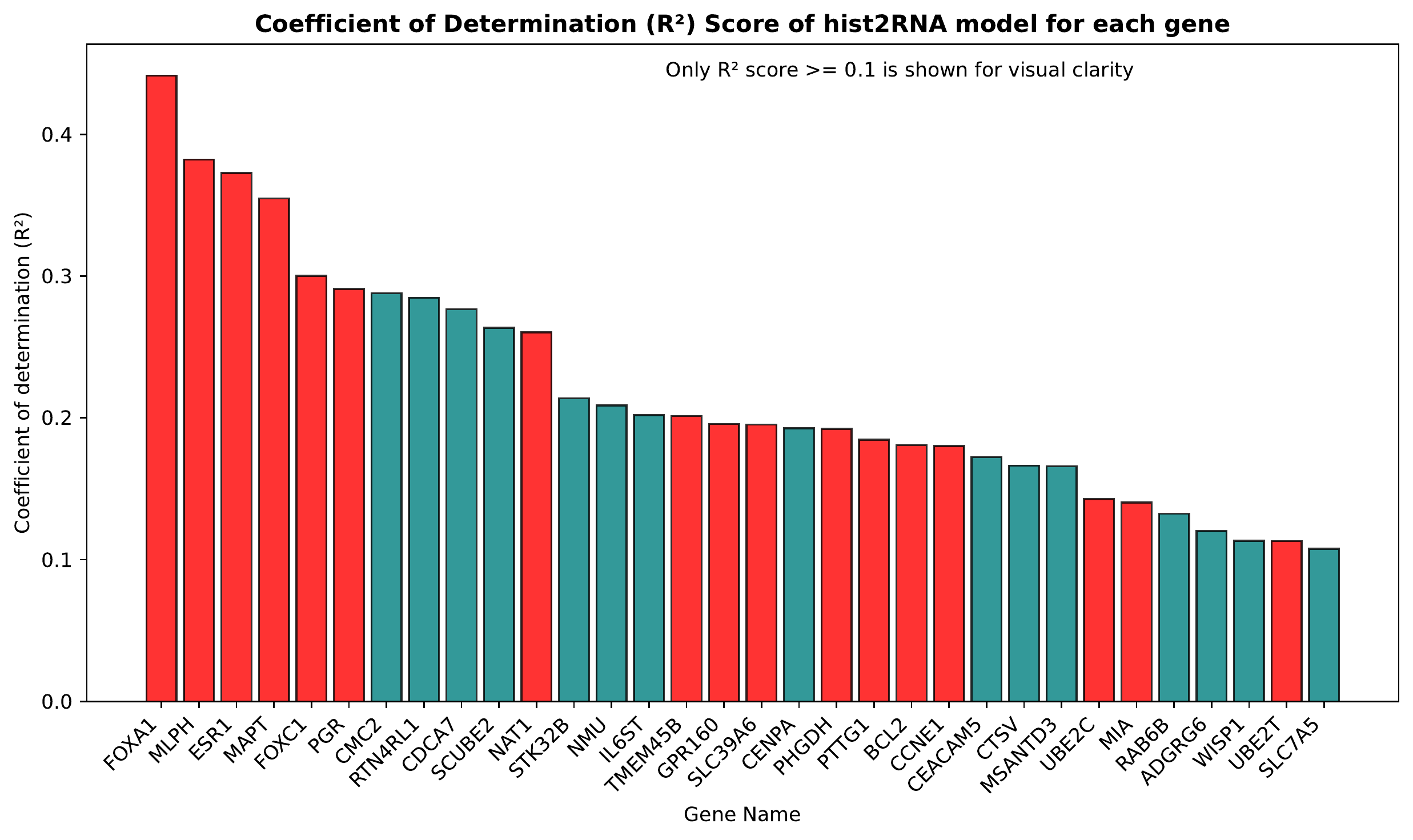}
\caption{Coefficient of determination (R$^2$) scores for the top 32 genes predicted by our model. The bar plot includes all genes with an R$^2$ score greater than or equal to 0.1. PAM50 genes, which are important biomarkers for breast cancer subtyping, are highlighted in red.}
\label{r2score}
\end{figure}

\begin{figure}[!t]
\centering
\includegraphics[width=0.65\textwidth]{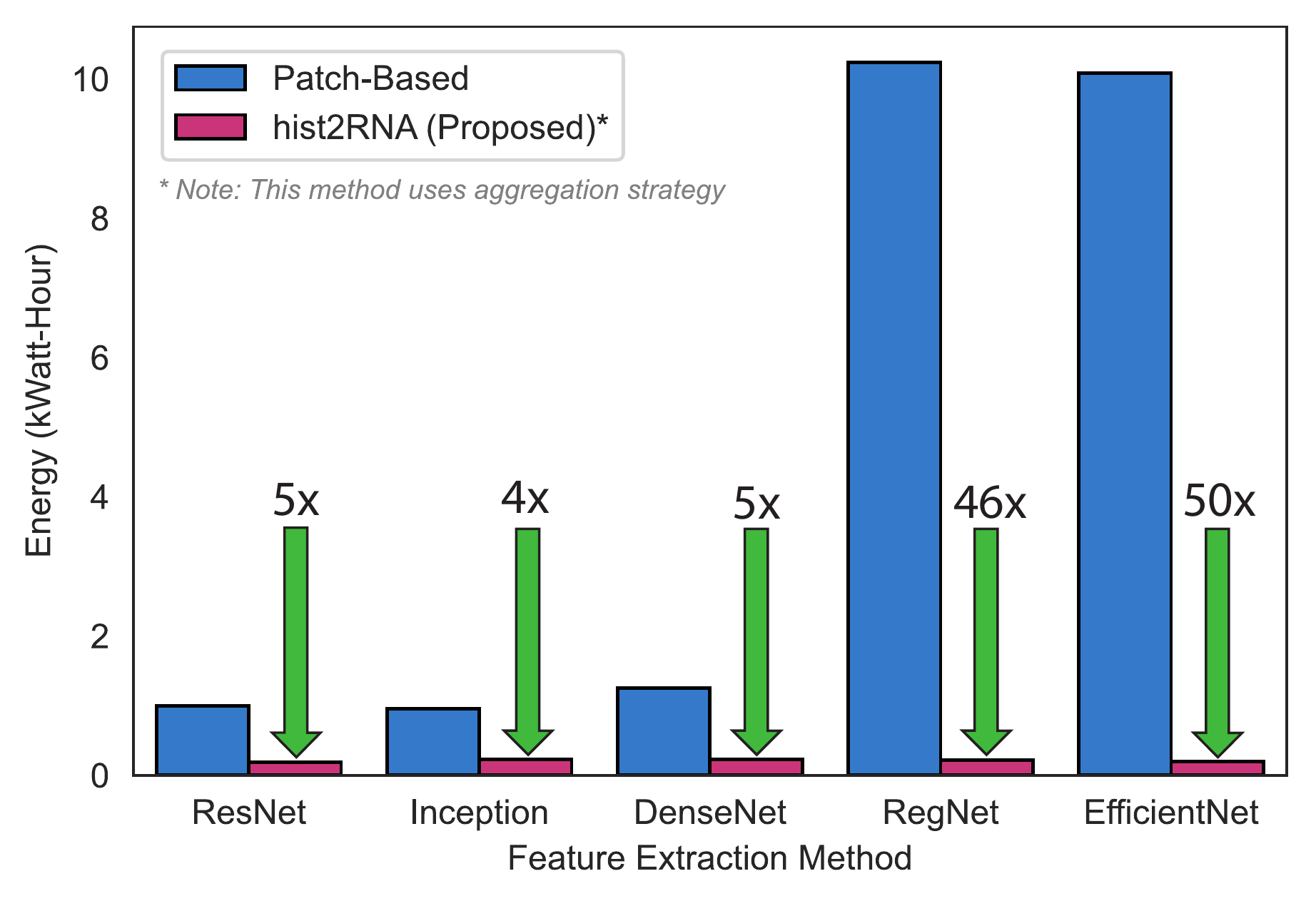}
\caption{Energy consumption comparison based on the training times of the proposed strategy versus a patch-based strategy. To calculate energy (in kilo-Watt-hours, kWh), the power consumption of the used NVIDIA Tesla V100 32GB GPUs (about 300 W each) was multiplied by the training time (in hours) and divided by 1,000. For example, training the EfficientNet patch-based method on four GPUs took 8.48 hours at maximum power, translating to an energy consumption of 4 $\times$ 8.48 $\times$ 300 $=$ 10.176 kWh. By contrast, training our EfficientNet aggregation-based approach took only 0.66 hours on a single GPU, thus consuming only 0.19 kWh, which is 50$\times$ lower than the patch-based method.}
\label{energy}
\end{figure}  

\begin{figure}[!t]
\includegraphics[width=16.5cm]{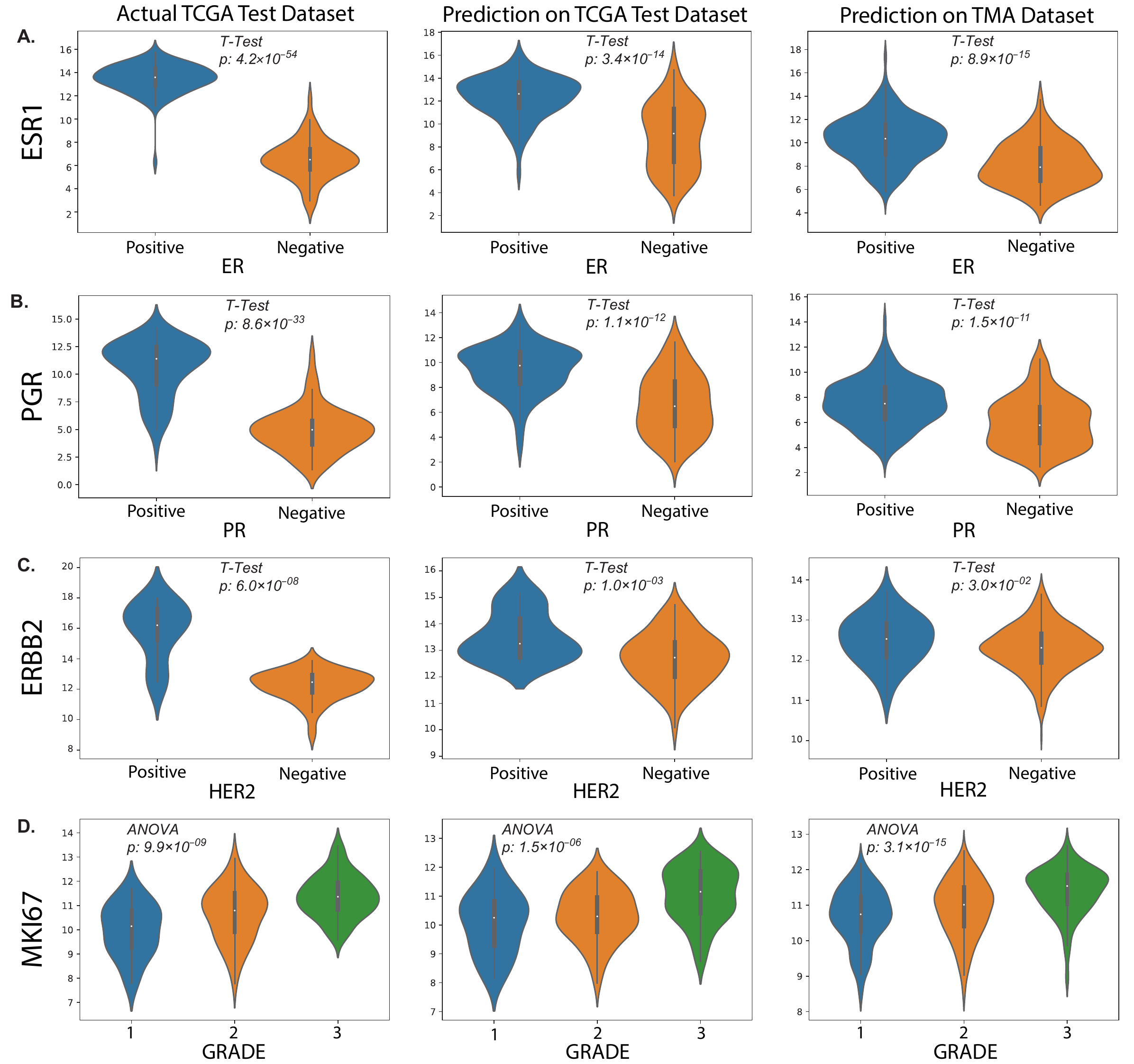}
\caption{Predicted results from our proposed method. The plots show the distributions of RNA-sequencing expression values for routine biomarkers ESR1 (A), PGR (B), ERBB2 (C) and MKI67 (D) according to the TCGA dataset (left column) and their predicted values on the held-out test set (middle column) and our external TMA dataset (right column), with respect to clinical status (IHC) of protein expression for the corresponding proteins encoded by each gene.}
\label{violinplot}
\end{figure}  

\begin{figure}[!t]
\centering
\includegraphics[width=0.7\textwidth]{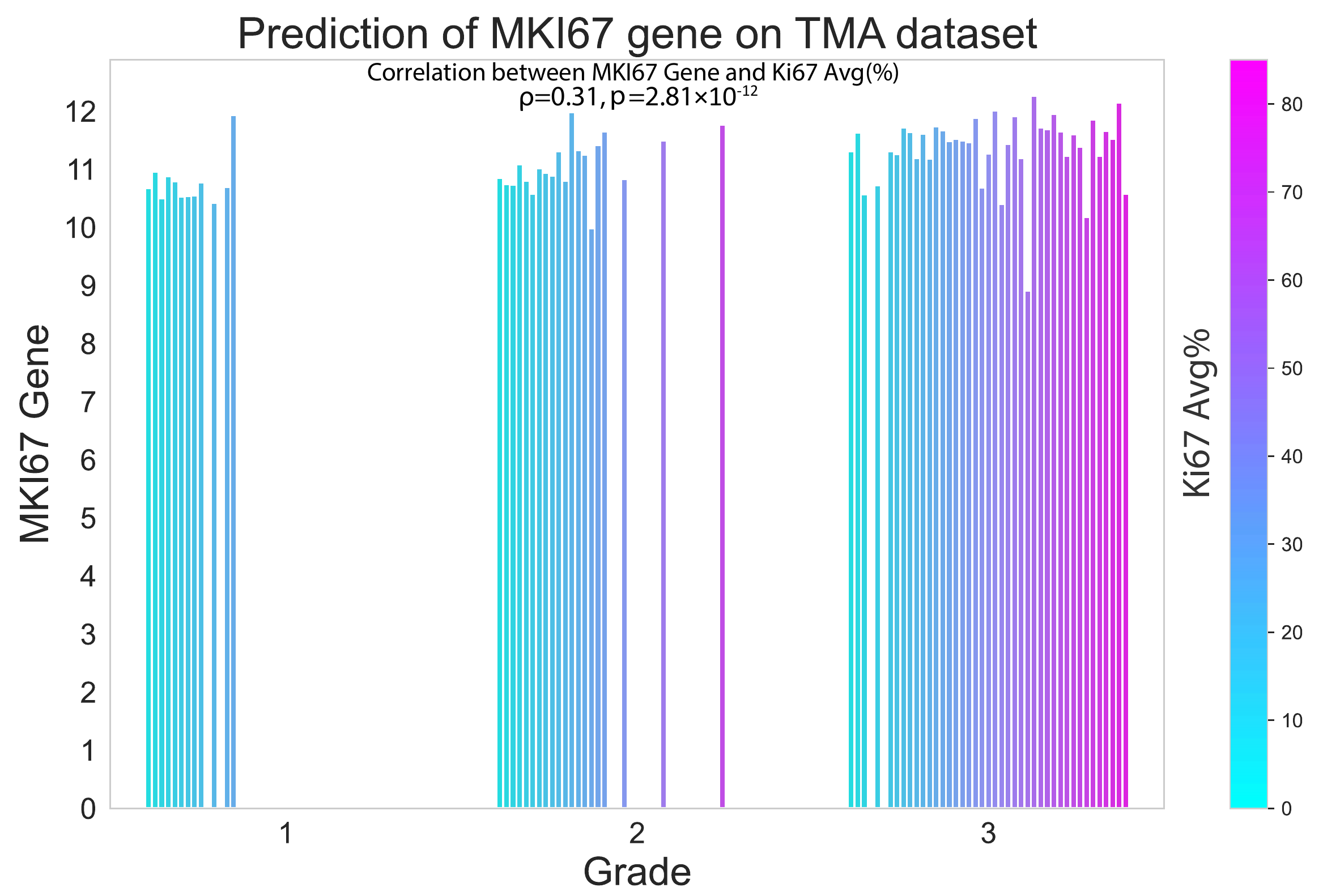}
\caption{Proposed model performance in predicting MKI67 gene expression on the external TMA dataset. The results are seen to be significantly correlated with the Ki67\% value.}
\label{mik67}
\end{figure}  
\section{Results}
\subsection{Evaluating hist2RNA Method on Held-Out Test Data}
We evaluated the proposed hist2RNA method (Figure~\ref{fig:architecture}) with five different pretrained models across patients and genes (Figure~\ref{box_plot}). Of the five models, EfficientNet appeared favourable, achieving a median Spearman correlation of 0.82 across patients on the held-out test set with few outliers, similar to the other models, while achieving the highest median correlation of 0.29 across genes, with statistically significant results for 105 out of 138 genes (\emph{p} < 5 $\times$ 10\textsuperscript{$-$2}). Of the top 20 genes (Figure~\ref{scatter_plot}) according to their corresponding correlation coefficients, 13 belong to the PAM50 gene set. In addition, 32 genes predicted by our model had a coefficient of determination (R$^2$) $\geq$ 0.1 (Figure~\ref{r2score}), of which 17 were from PAM50.

\subsection{Classification of Breast Cancer Subtypes and Computational Efficiency of hist2RNA}
Next, we aimed to develop a gene expression-based classification model for breast cancer subtypes using the hist2RNA-predicted PAM50 gene set. We used a voting classifier consisting of four different machine learning algorithms, including Random Forests (RF), MultiLayer Perceptron (MLP), Linear Discriminant Analysis (LDA), and Logistic Regression (LR), with soft voting. In this experiment, the hist2RNA-predicted PAM50 gene set achieved reasonable accuracy and F1 score of 0.55 and 0.54, respectively, for the classification of the four subtypes (LumA, LumB, Basal and HER2). Our model achieved an AUROC (Area Under the Receiver Operating Characteristic curve) score of 0.79, 0.63, 0.89, and 0.78 for LumA, LumB, Basal and HER2 subtypes. In terms of computational efficiency, the proposed method outperformed the patch-based approach as evidenced by the results (Figure~\ref{energy}), which showed $50\times$ lower energy consumption, which can be attributed to EfficientNet's reduced training time.

{\subsection{Performance of hist2RNA on External TMA Dataset}
In this study, we also evaluated the performance of our proposed method on an external TMA dataset. The results presented in the violin plot (Figure~\ref{violinplot}) indicate that there were statistically significant differences in the predicted median gene expression of ESR1, PGR and ERBB2 for ER, PR and HER2 positive and negative cases respectively, in both TCGA and TMA datasets. Similarly, median gene expression of MKI67 showed statistically significant differences across the three grades correlated with increasing proliferation and biological aggressiveness. In addition, our analysis revealed a positive correlation ($\rho$ = 0.31, p = 2.81 $\times$ 10\textsuperscript{$-$12}) between predicted MKI67 gene and average Ki67\% protein expression, as illustrated in Figure~\ref{mik67}. It is worth noting that higher cancer grades are often associated with higher Ki67\% values, which is confirmed by this study.

As mRNA gene expression data was not available for the external TMA dataset, we instead evaluated the surrogate subtype classification based on clinically used IHC-based methods. In order to do this, we reclassified the PAM50 subtype using predicted genes from the TMA dataset, and first conducted patient-level univariate survival analysis solely using the predicted subtypes. The results revealed a c-index of 0.56, with an HR of 2.16 (95\% CI 1.12-3.06), and \emph{p} < 5 $\times$ 10\textsuperscript{$-$3} (Figure~\ref{survival} and Table~\ref{hazard-table1}). Subsequently, we performed multivariate analysis by incorporating standard clinicopathological variables such as tumour grade, tumour size, age and lymph node status. The subtype prediction demonstrated independent prognostic significance with a c-index of 0.65, HR = 1.87 (95\% CI 1.30-2.68), and \emph{p} < 5 $\times$ 10\textsuperscript{$-$3} (Table \ref{hazard-table1}). The p-values for both methods were statistically significant, indicating that both methods are able to distinguish between LumB and LumA subtypes with a high degree of confidence. IHC-based classification resulted in 96 patients being classified as LumB and 309 patients as LumA. In contrast, the proposed hist2RNA model resulted in 69 patients being classified as LumB and 336 patients being classified as LumA. This represents a shift of 27 patients from LumB to LumA in the proposed method compared to IHC-based classification. With regard to other clinopathological parameters, tumour size and patient age were consistently observed to be significant predictors of overall survival among luminal breast cancer patients, with HRs of around 1.6 and 3 respectively in both univariate and multivariate analyses (Table~\ref{hazard-table1} and \ref{hazard-table2}), where \emph{p} < 5 $\times$ 10\textsuperscript{$-$3}. Despite being a commonly used prognostic factor, tumour grade was not observed as a significant predictor, as indicated by \emph{p} > 5 $\times$ 10\textsuperscript{$-$2} in both univariate and multivariate analyses.
  
\section{Discussion}
\subsection{hist2RNA Model Performance in Gene Expression Prediction}
We have demonstrated reliable and robust RNA-sequencing gene expression prediction from breast cancer WSIs with our proposed hist2RNA model, which was validated on a held-out test set from TCGA. The results suggest that our model using the EfficientNet pretrained model for image feature extraction is well-suited for predicting gene expression across a diverse set of 138 genes related to risk prediction in ER+ breast cancer. The proposed aggregation strategy achieved a median Spearman correlation of 0.29 (FDR adjusted \emph{p} = 3.8 $\times$ 10\textsuperscript{$-$4}) across genes and 0.82 (FDR-adjusted \emph{p} = 4.3 $\times$ 10\textsuperscript{$-$64}) across patients, while also being computationally efficient in terms of energy consumption compared to the patch-based approach. The findings of the study reveal that the correlation coefficient for gene prediction across patients was significantly higher (0.82) in comparison to the correlation coefficient for gene prediction across 138 genes (0.29). This outcome can be attributed to the fact that gene prediction across patients takes into account the overall patterns of gene expression for each patient, allowing the model to capture the general trends in gene expression that are shared across patients. Conversely, when evaluating gene prediction across 138 genes, the correlation coefficient is calculated for each gene separately. This approach is more focussed on the individual gene expression patterns which can exhibit significant variation between patients, thereby resulting in lower correlation coefficients.

\subsection{Comparing Breast Cancer Subtype Classification Approaches}
In this study, we carried out breast cancer subtype classification by leveraging the PAM50 gene set predicted from hist2RNA-based models. When comparing breast cancer subtype results with other studies, it is important to first acknowledge the differences in approach. Both Kather et al. and Liu et al. utilised direct image-to-subtype classification techniques, whereas our proposed method involves a two-step process: first image-to-gene prediction, followed by gene-to-subtype classification~\cite{Kather2020}~\cite{liu22}. Despite these differences, our study found that the PAM50 gene set predicted by hist2RNA outperformed Kather et al.'s results in terms of AUROC scores across all subtypes: LumA (0.79 vs 0.78), LumB (0.63 vs 0.61), Basal (0.89 vs 0.85), and HER2 (0.78 vs 0.75). Furthermore, our study achieved an accuracy of 56\% and an F1 score of 55\% in breast cancer subtypes classification, while Liu et al. achieved higher accuracy of 64.3\% and higher F1 score of 68.5\%. Our comparison of results with Liu et al. suggest that direct image-to-subtype approach can yield good performance in breast subtype classification. Therefore, further optimisation of gene expression generation process may be necessary to enhance the classification performance.

\begin{table}[!t]
\caption{Univariate and multivariate analysis for overall survival in the TMA dataset of Luminal patients using hist2RNA predicted genes.}
\label{hazard-table1}
\resizebox{\columnwidth}{!}{%
\begin{tabular}{llllllllll}
\hline
\multirow{2}{*}{\textbf{Parameter}} & \multirow{2}{*}{\textbf{\begin{tabular}[c]{@{}l@{}}Risk group\\ cut-off value\end{tabular}}} & \multirow{2}{*}{\textbf{\begin{tabular}[c]{@{}l@{}}No. of patients\\ in each group\end{tabular}}} & \multicolumn{3}{l}{\textbf{Multivariate (n=406)}} & & \multicolumn{3}{l}{\textbf{Univariate (n=406)}} \\
\cline{4-6}\cline{8-10}
& & & \textbf{HR} & \textbf{95\% CI} & \textbf{p} & & \textbf{HR} & \textbf{95\% CI} & \textbf{p} \\
\hline
Tumour Grade & 1 \& 2 vs. 3 & 324 vs. 81 & 0.94 & 0.62-1.42 & 7.6 $\times$ 10\textsuperscript{$-1$} & & 0.92 & 0.31-1.36 & 6.7 $\times$ 10\textsuperscript{$-1$} \\
Tumour Size & >20  vs. $\leq20$ (mm) & 297 vs. 108 & 1.57 & 1.11-2.21 & 1 $\times$ 10\textsuperscript{$-2$} & & 1.56 & 0.77-2.15 & 1 $\times$ 10\textsuperscript{$-2$} \\
Age & >55 vs. $\leq55$ & 139 vs. 266 & 3.04 & 2.02-4.58 & <5 $\times$ 10\textsuperscript{$-3$} & & 2.96 & 1.49-4.43 & <5 $\times$ 10\textsuperscript{$-3$} \\
LN status & pos. vs. neg. & 283 vs. 122 & 1.24 & 0.89-1.73 & 2 $\times$ 10\textsuperscript{-1} & & 1.43 & 0.67-1.96 & 3 $\times$ 10\textsuperscript{-2} \\
hist2RNA Predicted & LumB vs. LumA & 69 vs. 336 & 1.87 & 1.30-2.68 & <5 $\times$ 10\textsuperscript{-3} & & 2.16 & 1.12-3.06 & <5 $\times$ 10\textsuperscript{$-3$} \\
\hline
\end{tabular}%
}
\end{table}

\begin{table}[!t]
\caption{Univariate and multivariate analysis for overall survival in the TMA dataset of Luminal patients using IHC based subtype.}
\label{hazard-table2}
\resizebox{\columnwidth}{!}{%
\begin{tabular}{llllllllll}
\hline
\multirow{2}{*}{\textbf{Parameter}} & \multirow{2}{*}{\textbf{\begin{tabular}[c]{@{}l@{}}Risk group\\ cut-off value\end{tabular}}} & \multirow{2}{*}{\textbf{\begin{tabular}[c]{@{}l@{}}No. of patients \\ in each group\end{tabular}}} & \multicolumn{3}{l}{\textbf{Multivariate (n=406)}} & & \multicolumn{3}{l}{\textbf{Univariate (n=406)}} \\
\cline{4-6}\cline{8-10}
& & & \textbf{HR} & \textbf{95\% CI} & \textbf{p} & & \textbf{HR} & \textbf{95\% CI} & \textbf{p} \\
\hline
Tumour Grade & 1 \& 2 vs. 3 & 324 vs. 81 & 0.72 & 0.45-1.13 & 1.5 $\times$ 10\textsuperscript{$-$1} & & 0.92 & 0.31-1.36 & 6.7 $\times$ 10\textsuperscript{$-1$} \\
Tumour Size & >20 vs. $\leq20$ (mm) & 297 vs. 108 & 1.63 & 1.16-2.30 & <5 $\times$ 10\textsuperscript{$-$3} & & 1.56 & 0.77-2.15 & 1 $\times$ 10\textsuperscript{$-$2} \\
Age & >55 vs. $\leq55$ & 139 vs. 266 & 3.21 & 2.13-4.83 & <5 $\times$ 10\textsuperscript{$-$3} & & 2.96 & 1.49-4.43 & <5 $\times$ 10\textsuperscript{$-$3} \\
LN status & pos. vs. neg. & 283 vs. 122 & 1.27 & 0.91-1.77 & 2 $\times$ 10\textsuperscript{$-1$} & & 1.43 & 0.67-1.96 & 3 $\times$ 10\textsuperscript{$-$2} \\
IHC Subtype & LumB vs. LumA & 96 vs. 309 & 2.07 & 1.42-3.02 & <5 $\times$ 10\textsuperscript{$-$3} & & 1.68 & 1.20-2.34 & <5 $\times$ 10\textsuperscript{$-$3} \\
\hline
\end{tabular}%
}
\end{table}

\begin{figure}[!t]
\centering
\includegraphics[width=\textwidth]{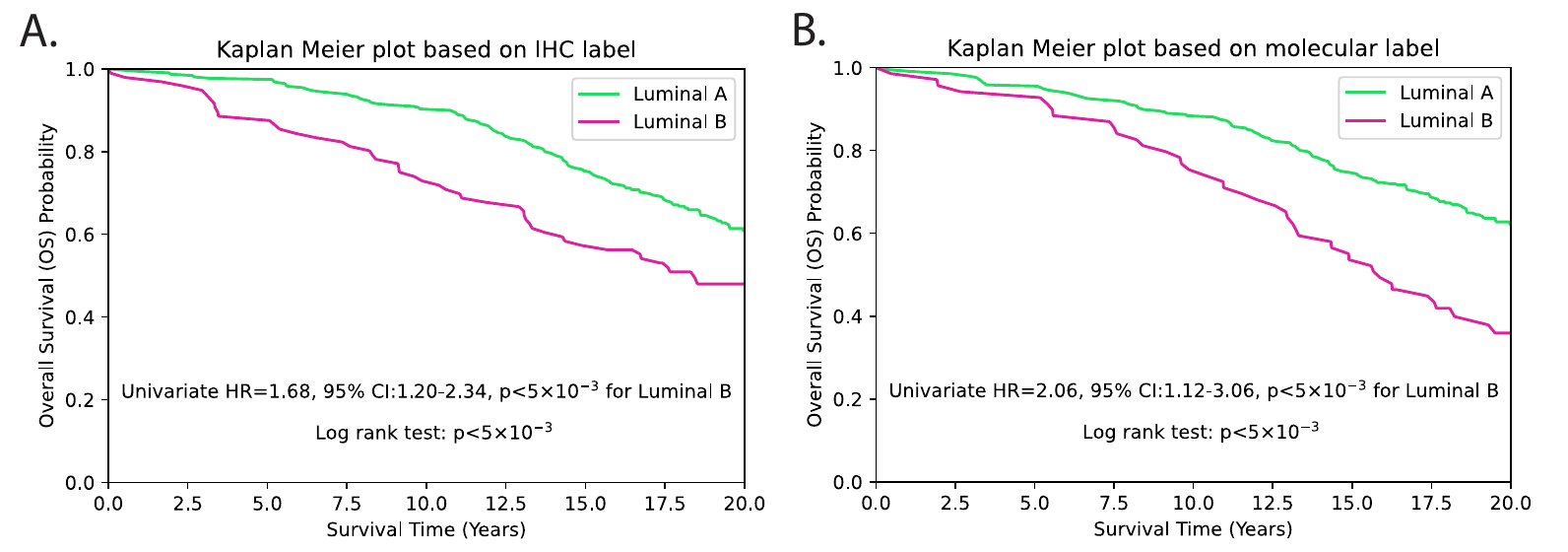}
\caption{Kaplan–Meier survival estimation of luminal (LumA/LumB) categories on the TMA dataset using the IHC (A) and molecular (B) subtype label from hist2RNA predicted genes. The log-rank test p-value indicates a significant difference in survival between the two subtypes (\emph{p} < 5 $\times$ 10\textsuperscript{$-$3}).}
\label{survival}
\end{figure}
\subsection{Clinical Implications of hist2RNA Model in Breast Cancer}
We applied our findings to an independent external cohort of patients from a randomised clinical trial to demonstrate that our model has direct prognostic significance. The PAM50 gene set was chosen to predict subtypes such as LumA and LumB, and to evaluate its potential impact on improving patient survival. Furthermore, the clinical use of Ki67 as a proliferation marker remains standard practice in many countries due to the high cost of molecular profiling. The positive correlation (0.31) between our predicted MKI67 gene and Ki67\% value by IHC staining suggests that our proposed method can serve as an alternative to IHC. We also showed that higher MKI67 expression value is associated with higher grade cancer and found statistically significant differences with lower grade cancer in both test and external data. Our results suggest that predicting ESR1 and PGR gene expression from WSIs may potentially be a good method for differentiating ER and PR protein status in breast cancer patients. The significant differences observed in the predicted gene expression levels between positive and negative cases, along with the strong positive correlations found for ESR1 ($\rho$ = 0.57, \emph{p} = 1.87 $\times$ 10\textsuperscript{$-$15}) and PGR ($\rho$ = 0.54, \emph{p} = 1.52 $\times$ 10\textsuperscript{$-$13}) in the test set, supports this conclusion. However, based on our findings, the association between ERBB2 gene and HER2 status was not as strong as that observed for ESR1 and PGR, which may be due to the complex regulation of HER2 expression, the influence of other genetic factors or the relatively small sample size of our dataset.

\subsection{Multigene Prognostic Signatures and Risk Stratification}
Our multigene prediction model generates crucial prognostic information particularly useful for cancer patients where clinical parameters and traditional IHC markers alone lead to equivocal findings. The clinical importance of multigene prognostic signatures in cancer is extensively demonstrated in the current clinical practice guidelines for adjuvant systemic therapy in early-stage breast cancer and the disease staging guidelines \cite{Vieira2018}. In this study, our model reclassified about 28\% of LumB patients as LumA, with corresponding good outcomes. This shift in classification could be due to the inclusion of molecular information that is not captured by IHC-based classification. This additional information may better differentiate between the LumB and LumA subtypes and result in more accurate risk stratification. A closer analysis of the HRs and CIs suggests that the proposed hist2RNA method may be a slightly better predictor of LumB subtype compared to the IHC subtype method. Univariate analysis of the proposed method showed higher risk of death for LumB compared to LumA subtypes, with a hazard ratio (HR) of 2.16 (95\% CI 1.12-3.06, \emph{p} < 5 $\times$ 10\textsuperscript{$-$3}), while the IHC method had a lower HR of 1.68 (95\% CI 1.20-2.34, \emph{p} < 5 $\times$ 10\textsuperscript{$-$3}) for the same subtypes (see Table \ref{hazard-table2}). In contrast, multivariate analysis of LumB versus LumA subtypes indicated that the proposed method yielded a hazard ratio of 1.87 (95\% CI 1.30-2.68), whereas the IHC method showed a slightly higher hazard ratio of 2.07 (95\% CI 1.42-3.02). However, the CI for the proposed method was narrower, suggesting greater precision in the estimate. This suggests that the proposed model may be more accurate in predicting patient outcomes and risk stratification compared to the IHC-based classification, particularly in identifying patients with the LumB subtype who are at higher risk of recurrence and worse outcomes. The observed shift of 27 patients from LumB to LumA in our study suggests less aggressive treatment recommendations and potential avoidance of chemotherapy toxicities. Overall, the findings of this study are encouraging, and we anticipate that our method may be useful for predicting other cancer types as well as other molecular phenotypes, such as mutation status prediction, copy-number alterations and microbiome prediction. Prediction of molecular phenotypes can enable cost-effective precision medicine that will allow large-scale epidemiological studies that include gene expression phenotypes as exposures in the research domain \cite{Schmauch2020}. 

\subsection{Comparison with Other State-of-the-Art Approaches}
In comparing our proposed method with other state-of-the-art approaches, it is important to consider potential challenges arising from variations in datasets, target variables (number of genes) and evaluation metrics used in previous studies. Therefore, a direct comparison between our method and previous studies may not be feasible. For example, Weitza et al. utilised ResNet18 to extract prostate cancer features and applied hierarchical clustering to reduce the number of CNN features, achieving Spearman correlation of 0.243 with associated FDR adjusted \emph{p} < 1 $\times$ 10\textsuperscript{$-$4}~\cite{Weitza}. Schmauch et al. used a pan-cancer dataset from TCGA and employed ResNet50 to extract features, followed by k-means clustering to reduce dimensionality~\cite{Schmauch2020}. They achieved tile-level Pearson correlation of 0.19 on T-Cell genes and 0.23 on B-cell genes, with \emph{p} < 1 $\times$ 10\textsuperscript{$-$4} in both cases. Tavolara et al. utilised attention-based multiple instance learning on mouse population data to predict the expression of the top five frequently occurring genes, achieving an average Pearson correlation coefficient of 0.59~\cite{Tavolara2021}. Lastly, Wang et al. utilised a patch-based strategy and achieved a predicted median Spearman correlation coefficient of 0.4 with Bonferroni adjusted \emph{p} < 5 $\times$ 10\textsuperscript{$-$2}. However, their approach took 12-70 hours on a single GPU (NVIDIA Tesla V100 32GB) for each gene~\cite{Wang2021f}. In contrast, our experiments indicated that the patch-based approach inspired by Wang et al.'s method yields a median Spearman correlation coefficient of 0.18 using RegNet on our test set, with 68 genes having \emph{p} < 5 $\times$ 10\textsuperscript{$-$2} (Figure~\ref{box_plot_patch_based}).
\begin{figure}[!t]
\includegraphics[width=\textwidth]{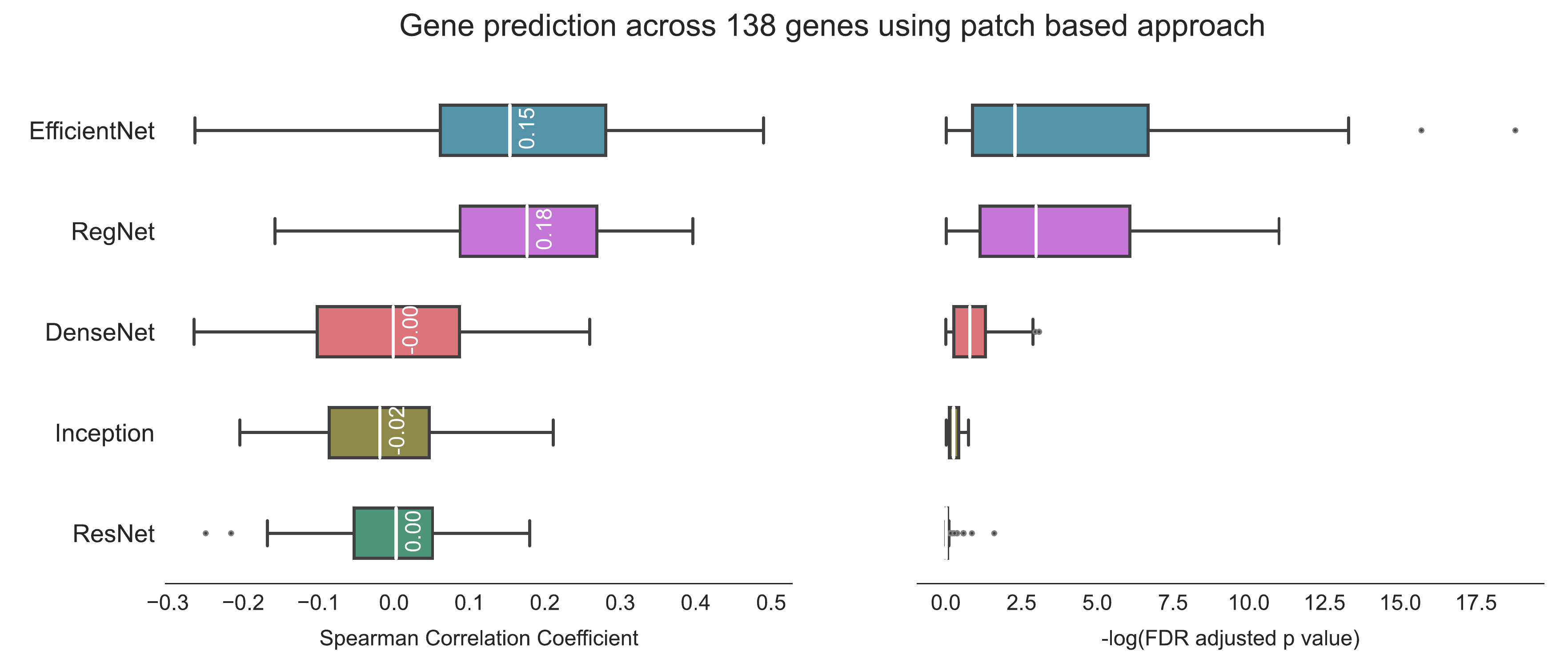}
\caption{Gene prediction across 138 genes using the patch-based approach \cite{Wang2021f}. The best result with this approach is a median correlation coefficient of 0.18 using RegNet on our test data, with the corresponding $-$log FDR adjusted p-value indicating statistical significance (where significance is represented by $-$log(5\% FDR p value) > 1.30). Notably, higher $-$log FDR adjusted p-values are considered to be indicative of better model performance.}
\label{box_plot_patch_based}
\end{figure}  
\subsection{Limitations, Challenges and Future Directions}
Despite promising results, our study has some limitations. Generating gene expression data from images has the potential to introduce extra noise into the subtype classification task due to factors such as tissue heterogeneity, staining variability and image artefacts. Although we used an external dataset and performed exploratory analysis, we did not perform rigorous external validation of our method due to the absence of molecular information in the TMA dataset. Additionally, the external dataset we used had limited representation of the Basal and HER2 subtypes. Therefore, our exploratory analysis was limited to the LumA and LumB subtypes (n=405). This limitation may affect the generalisability of our findings to other breast cancer subtypes, and further validation on a more diverse dataset is warranted. Six other studies have reported on predicting gene expression from histopathology images, each with its own limitations, including small sample size, significant training time, individual gene-based training, lack of independent external validation cohorts and lack of comparison with other pretrained models suggesting generally that more work is necessary~\cite{Hong2021}\cite{Schmauch2020}\cite{Schrammen2021}\cite{Weitza}\cite{Wang2021f}\cite{Tavolara2021}.

Currently, it would be premature to propose that image analysis can replace gene expression assays for clinical applications. While there has been promising research on the use of image analysis to predict gene expression, several challenges need to be addressed before it can be implemented in clinical practice. These challenges include the development of robust and reliable algorithms for image analysis, and the validation of image-based biomarkers in large and diverse patient populations. Finally, regulatory approval and integration of image analysis into clinical practice is needed, which requires collaboration between researchers, clinicians and regulatory agencies.

\section{Conclusions}
Summarising our findings, we conclude that mRNA expression in solid tumours can be effectively inferred from routine histology alone using deep learning approaches. We employed a public dataset and validated model robustness on a held-out test set, as well as conducting exploratory analysis on an external TMA dataset. Gene prediction from pathology images has substantial impact on patient survival by accurately identifying their molecular subtypes. These findings have the potential to pave the way for more personalised treatment planning for breast cancer patients in a time-efficient and cost-effective manner.

\section*{Data Availability}
The TMA image dataset and clinical data are not publicly available due to ethics restrictions, however they  may be accessible on reasonable request to the corresponding author. TCGA image data and clinical data are available publicly through \url{https://portal.gdc.cancer.gov/}(Accessed August 20, 2021).

\section*{Code Availability}

Our work is fully reproducible and source code is available at \href{https://github.com/raktim-mondol/hist2RNA}{Github}.

\section*{Institutional Review Board Statement} 
Ethical approval was provided by the South Eastern Sydney Local Health District Human Research Ethics Committee at Prince of Wales Hospital (HREC 96/16). All patients recruited to the trial were consented. All methods were performed in accordance with the relevant institutional guidelines and regulations.

\section*{Informed Consent Statement}
Informed consent was obtained from all subjects involved in the study.


\section*{Acknowledgement}
EKAM is supported by a Researcher Exchange and Development within Industry (REDI) Fellowship from MTPConnect \& MRFF Australia. This research was undertaken with the assistance of resources and services from the National Computational Infrastructure (NCI), which is supported by the Australian Government. Additionally, data preprocessing was performed using the computational cluster Katana, which is supported by Research Technology Services at UNSW Sydney.




\bibliographystyle{ieeetr}  
\bibliography{references}

%


\end{document}